\documentclass[letterpaper, 10 pt, journal, twoside]{IEEEtran}
\usepackage{amsmath,amsfonts}
\usepackage{amssymb}
\usepackage{algorithmic}
\usepackage{array}
\usepackage{caption}
\usepackage{textcomp}
\usepackage{stfloats}
\usepackage{url}
\usepackage{gensymb}
\usepackage{hyperref}
\usepackage{verbatim}
\usepackage{graphicx}
\usepackage{xcolor,colortbl}
\usepackage{tabularx,booktabs}
\usepackage{array,multirow,graphicx}
\usepackage{makecell}
\usepackage[export]{adjustbox}
\usepackage{pifont} 
\usepackage{etoolbox}
\usepackage[normalem]{ulem}
\usepackage{booktabs}
\usepackage{threeparttable}
\usepackage{tabularx}
\usepackage{multirow}
\usepackage[printonlyused]{acronym}
\usepackage{siunitx}
\usepackage{subcaption}
\usepackage{ragged2e}
\usepackage{subcaption}

\usepackage{enumitem} 
\usepackage[dvipsnames]{xcolor}

\colorlet{colorFst}{Green!25}       %
\colorlet{colorSnd}{SpringGreen!45} %
\colorlet{colorTrd}{Yellow!30}      %

\hypersetup{
    colorlinks=true,
    urlcolor=magenta
}

\newacro{pHRI}{physical human-robot interaction}
\newacro{HRC}{Human-Robot Collaboration}
\newacro{PPO}{Proximal Policy Optimization}
\newacro{KL}{Kullback-Leibler}
\newacro{FT}{Force-Torque}
\newacro{LL}{Low-Level}
\newacro{HL}{High-Level}

\begin{document}


\author{Zhihao Cao$^{1, *}$, Tianxu An$^{1,2,*}$, Chenhao Li$^{1,3}$, Stelian Coros$^{2}$, Marco Hutter$^{1}$



\thanks{This work was supported by Hilti AG, Schaan, Liechtenstein, the Luxembourg National Research Fund (Ref. 18990533), the Swiss National Science Foundation (Ref. 200021E\_229503), and by the ETH AI Center.}

\thanks{$^{1}$ Robotic Systems Lab; ETH Zurich, Switzerland
}
\thanks{$^{2}$ Computational Robotics Lab; ETH Zurich, Switzerland
}
\thanks{$^{3}$ ETH AI Center; Switzerland
}

\thanks{* \textit{Zhihao Cao and Tianxu An contributed equally to this work.} (\textit{Corresponding author: Tianxu An}).}

}

\title{PAINT: Partner-Agnostic Intent-Aware Cooperative Transport with Legged Robots}

\maketitle

\begin{abstract}
Collaborative transport requires robots to infer partner intent through physical interaction while maintaining stable loco-manipulation. This becomes particularly challenging in complex environments, where interaction signals are difficult to capture and model. We present PAINT, a hierarchical learning framework for partner-agnostic intent-aware collaborative legged transport that represents partner intent as an explicit interaction wrench and recovers it from payload-coupled proprioceptive histories. PAINT decouples intent understanding from terrain-robust locomotion: A high-level policy uses the inferred interaction wrench for transport and reconstructs it through teacher-student training, while a low-level locomotion backbone ensures robust execution. This enables lightweight deployment without external force-torque sensing or payload tracking. Extensive simulation and real-world experiments demonstrate compliant cooperative transport across diverse terrains, payloads, and partners. Furthermore, we show that PAINT can reuse the same single-agent policy for decentralized multi-robot transport by mechanically combining team interactions into a common wrench space, and supports heterogeneous-team transport. Our results suggest that payload-coupled proprioceptive interaction provides a physically grounded interface for partner-agnostic intent-aware collaborative transport.
\end{abstract}

\begin{IEEEkeywords}
Legged Robots, Cooperating Robots, Reinforcement Learning, Human-Robot Collaboration.
\end{IEEEkeywords}


\begin{figure}[htpb]
  \centering
  \includegraphics[width=0.49\textwidth]{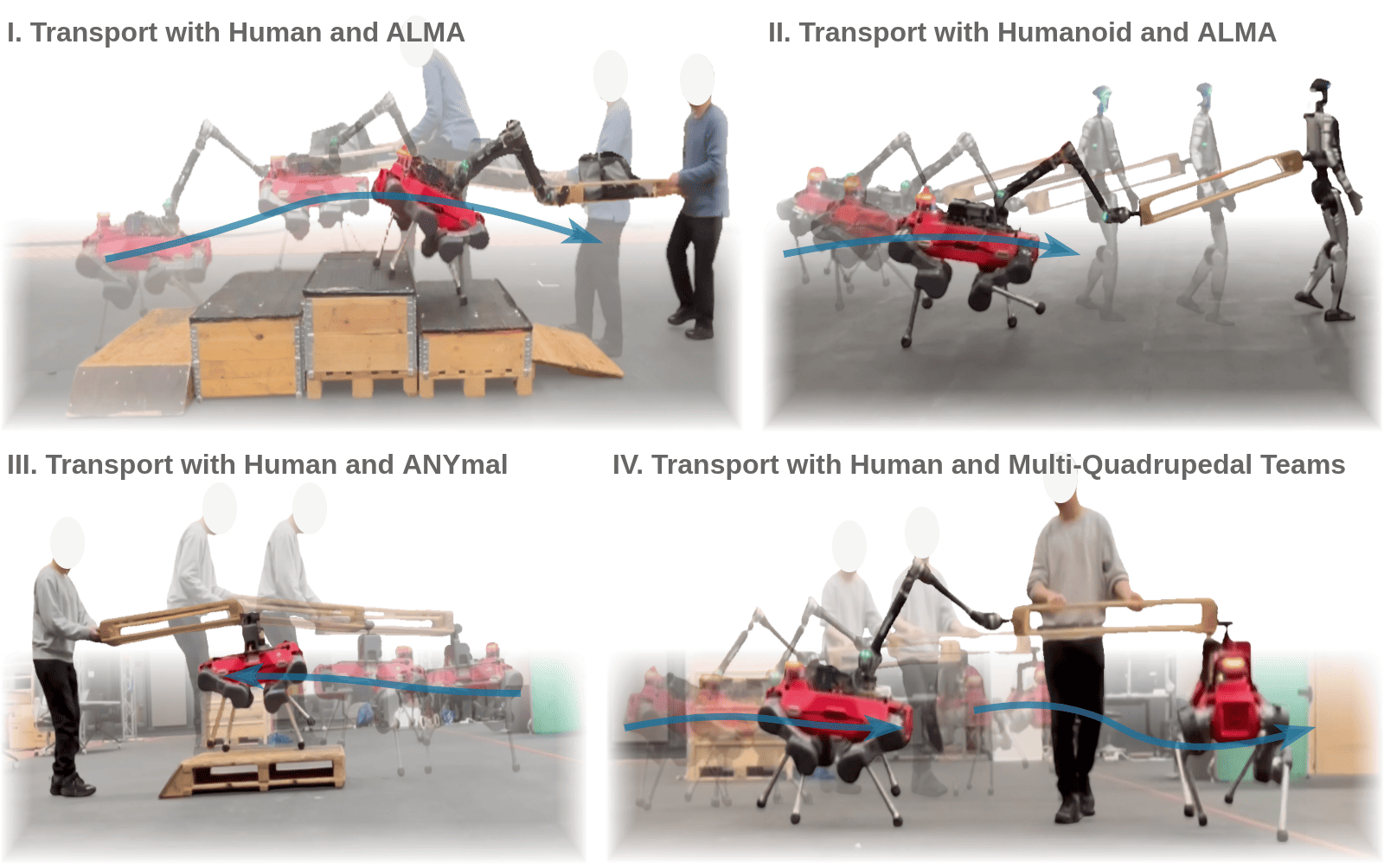}
  \vspace{-0.4\baselineskip}
  \caption{\textbf{Partner-agnostic intent-aware cooperative transport via purely proprioceptive feedback.} Our framework enables legged robots to transport a shared payload with diverse partners across various terrains. Relying solely on proprioceptive signals, the robot infers the partner’s intent without external force-torque sensors or explicit payload pose tracking during transport. This lightweight approach ensures robust, compliant coordination across heterogeneous transport team configurations and {can scale to decentralized multi-robot transport.} The project website is available at \url{https://paint-bot.github.io}.}
  \label{fig:method_overview}
  \vspace{-0.4\baselineskip}
\end{figure}

\section{Introduction}

Collaborative transport in heterogeneous teams of humans and robots is an important topic with applications in factory assistance and field operations. Collaborative robots can substantially reduce a partner's physical workload by assisting in the transport of large and heavy payloads \cite{li2024human}.
In practical collaborative transport, one partner typically acts as the leader, initiating the desired motion of the payload, while the collaborator follows by interpreting the leader's motion intent. 
Effective collaboration requires follower robots to maintain robust, terrain-aware locomotion under load while simultaneously inferring the partner's motion intent to achieve smooth, compliant transport \cite{du2025learning}.
In this work, we aim to develop a control pipeline that enables a robot to interpret and follow a leader’s motion intent during collaborative transport. Motion intent is represented by the force and torque that the leader applies to the payload to guide the transport movement \cite{yu2020human, bethala2025h, plotas2025control}. The leading partner may be either a human or another robot, and the transport team may include additional heterogeneous robots. Our goal is therefore to enable partner-agnostic collaboration during transport by inferring motion intent from the leading partner while maintaining robust locomotion over uneven terrains.

However, achieving this goal remains challenging for several reasons. First, motion intent is hard to capture and model across diverse environments with heterogeneous human-robot teams. For this reason, many intent-aware transport systems rely on external sensors, such as \ac{FT} sensors \cite{yu2020human, bethala2025h, plotas2025control}. Although informative, such sensors are often too expensive and fragile for high-impact deployment on factory floors, and are susceptible to locomotion-induced noise that is hard to filter out. Second, most existing intent-aware frameworks are validated only on flat laboratory terrain \cite{shao2024constraint, khandelwal2025compliant, gu2025hierarchical}. Their robustness on uneven terrain remains insufficiently explored. 
Finally, most previous studies focus on single human-robot transport pairs, while extending such interaction mechanisms to larger teams remains less explored. Existing multi-robot transport systems often rely on motion capture infrastructure \cite{zhang2026cognition}, remote control \cite{pandit2025multi}, or predefined team configurations \cite{de2023centralized, an2025collaborative}. Such assumptions limit portability to infrastructure-free deployment, particularly in outdoor settings, and make it harder to work with decentralized robotic teams or heterogeneous partners.

To this end, we propose \textbf{PAINT}, a lightweight hierarchical framework for \textbf{P}artner-\textbf{A}gnostic \textbf{INT}ent-aware cooperative transport using only accessible joint-level proprioceptive signals. {While PAINT's intent-aware formulation is not tied to a specific morphology and can be instantiated on different embodiments,} we specifically focus on quadrupedal robots for their robust locomotion on uneven terrains \cite{miki2022learning, he2025attention}.
PAINT decouples intent understanding from terrain-robust locomotion, where a \ac{HL} transport policy {infers partner intent as an interaction wrench recovered from proprioceptive histories} without \ac{FT} sensing or payload tracking, and a \ac{LL} locomotion backbone tracks base motion commands and stabilizes the robot in diverse uneven terrains.
We adopt a teacher-student learning scheme to make the learned \ac{HL} policy deployable under partial observability: a privileged teacher is trained with ground-truth interaction wrench data, while a student learns to reconstruct {the same interaction-wrench intent representation} via an intent estimator network from proprioception \cite{portela2024learning, zhi2025learning}. {This reconstruction is physically grounded in the payload-coupled dynamics: partner-applied interaction wrenches leave recoverable signatures in joint-level proprioceptive histories, enabling proprioception-only intent inference without direct wrench sensing. The same wrench space also supports decentralized multi-robot transport, where a rigid shared payload mechanically aggregates teammate interactions, allowing the same single-agent policy to be reused without multi-robot retraining.} The contributions of this work are as follows.

\begin{itemize}[itemsep=2pt,topsep=2pt,leftmargin=10pt]

    \item {\textbf{Payload-coupled proprioceptive intent inference.}
    We show that payload-coupled dynamics make partner-applied interaction wrenches identifiable from proprioceptive histories. This provides a physical basis for proprioception-only intent inference and enables the \ac{HL} estimator to recover intent without \ac{FT} sensing or payload tracking.}
    
    \item {\textbf{Lightweight hierarchical intent-aware transport policy.}
    We develop a lightweight hierarchical policy pipeline for intent-aware collaborative transport, where teacher-student training distills privileged wrench-conditioned behavior into a deployable proprioception-only \ac{HL} policy. A pretrained terrain-robust \ac{LL} locomotion backbone ensures stable transport over uneven terrains.}
    
    \item {\textbf{Payload-mediated decentralized multi-robot transport.}
    We demonstrate that the same single-agent policy can be reused for decentralized multi-robot transport, where the shared payload mechanically combines partner and teammate interactions into the same wrench space. This enables deployment without multi-robot retraining, and is validated in simulation and real-world transport.}
\end{itemize}
\begin{figure*}[htpb]
  \centering
  \includegraphics[width=0.98\textwidth]{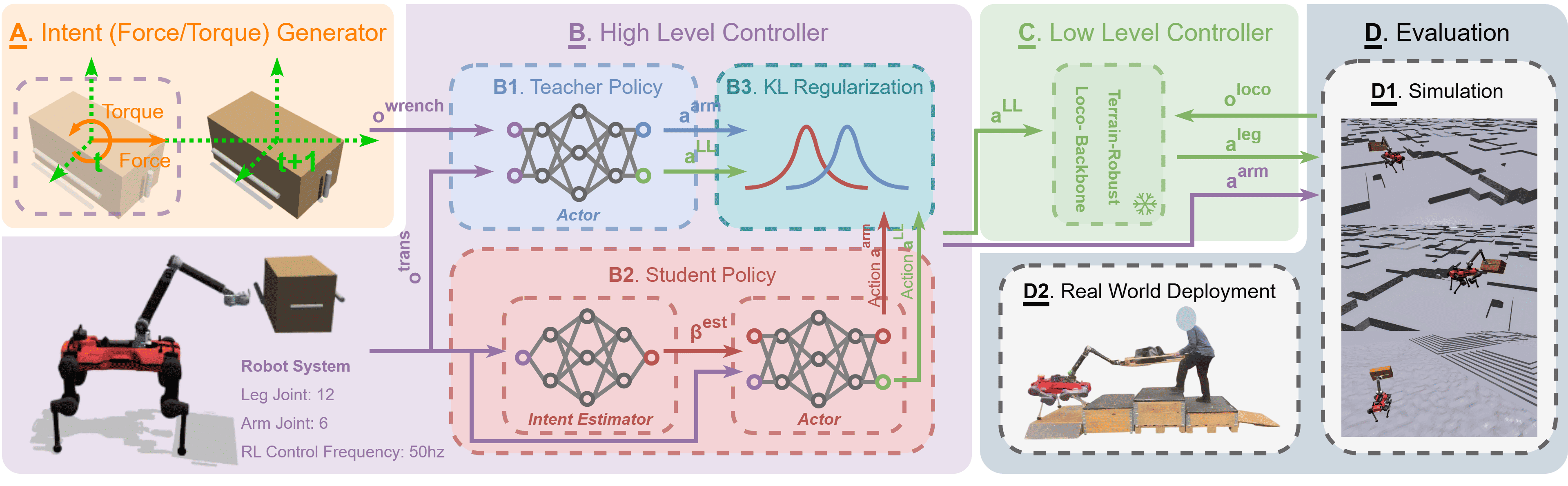}
  \vspace{-0.4\baselineskip}
  \caption{\textbf{Overview of the partner-agnostic intent-aware cooperative transport framework.} The system contains privileged training in simulation and lightweight real-world deployment via a hierarchical controller: \textbf{(A) Intent Generator:} Random forces and torques are applied to the payload to simulate partner guidance in simulation. \textbf{(B) High-Level Controller:} Trained with teacher-student scheme, a HL policy maps intent to base commands and arm joint actions, where a teacher \textbf{(B1)} uses wrench observations, and a deployable student \textbf{(B2)} relies on proprioceptive histories with an estimator to infer interaction wrench; KL regularization \textbf{(B3)} distills the teacher's action distribution into the student. \textbf{(C) Low-Level Controller:} A terrain-robust locomotion backbone tracks the HL commands for stable transport under load. \textbf{(D) Evaluation:} The framework is developed in simulation \textbf{(D1)} and transferred to real-world deployment \textbf{(D2)} without \ac{FT} sensing or payload tracking.}
  \label{fig:teaser}
  \vspace{-0.4\baselineskip}
\end{figure*}

\section{Related Work}

Cooperative transport is a classical case of \ac{HRC}. Early work emphasizes compliance via impedance-admittance regulation, enabling intuitive leader-follower behaviors and safe assistance \cite{ikeura1995variable}. Recent advancements are extending classical compliant cooperative transport toward intent-aware cooperative transport \cite{yu2020human, bethala2025h, plotas2025control, shao2024constraint}. Existing intent-aware approaches broadly follow two directions. Model-based methods achieve robustness through constraint-aware reasoning and structured inference \cite{yu2020human, shao2024constraint, khandelwal2025compliant, gu2025hierarchical, turrisi2024pacc}. They are effective under reliable state observability, but depend on accurate modeling. In contrast, learning-based intent encoding can better tolerate contact variability, adapt to complex payload-coupled dynamics, and enable richer coordination behaviors \cite{du2025learning, bethala2025h, pandit2025multi, an2025collaborative}. However, several systems still rely on \ac{FT} sensing \cite{yu2020human, bethala2025h, plotas2025control, schperberg2026safe}, which improves observability but increases deployment overhead due to cost and fragility. This has motivated lightweight alternatives that infer partner intent with only joint states and body dynamics \cite{du2025learning, shao2024constraint}, without \ac{FT} sensors at the end-effector. Beyond the sensing modality, {the form of the intent representation also matters. Structured physical representations, such as wrenches \cite{zhou2025hac}, can make the inferred intent task-aligned and interpretable, while providing an interface for downstream transport control.}

In addition, cooperative transport can build on progress in legged loco-manipulation, {which naturally addresses rough terrain \cite{du2025learning, turrisi2024pacc}. Whole-body frameworks \cite{portela2024learning, pan2025roboduet} coordinate arm and legs, but their end-to-end design can prevent reuse of established locomotion backbones \cite{miki2022learning, he2025attention, lee2020learning} and require costly retraining; hierarchical designs instead offer a simpler, more practical integration pathway \cite{ma2022combining, zhou2025hac}.}

Beyond single-robot \ac{HRC}, multi-robot cooperative transport introduces additional sensing and coordination challenges. Prior work stabilizes collaboration via centralized tracking and planning \cite{an2025collaborative}, motion capture \cite{zhang2026cognition}, remote control \cite{pandit2025multi}, or fixed team configurations \cite{de2023centralized, an2025collaborative, wang2025shared}, limiting scalability and field deployment. 
In addition, limited attention has been paid to inferring the team-level intent from physical payload-coupled interactions during multi-robot cooperative transport \cite{du2025learning}. 
In summary, prior cooperative-transport systems have not fully provided a deployable intent interface that is task-aligned, proprioception-only, and compatible with decentralized team coordination. PAINT addresses this gap by recovering a wrench-structured intent representation from payload-coupled proprioceptive histories and using it for high-level transport control. The same payload-induced wrench space further explains how a single-agent policy can be reused for decentralized multi-robot transport without multi-robot retraining.

\section{Method}
\label{sec:method}

\subsection{Overview}
\label{sec:method_overview}
We formulate partner-agnostic intent-aware cooperative transport under physical coupling as a Partially Observable Markov Decision Process (POMDP) $\mathcal{M}=(\mathcal{S},\mathcal{A},\mathcal{O},\mathcal{T},\mathcal{R},\Omega,\gamma)$.
The latent state $s_t\in\mathcal{S}$ is the coupled robot-payload-partner dynamics, which is only partially observable. At each step $t$, the agent receives $o_t\sim\Omega(o_t|s_t)$, takes action $a_t$, transitions via $\mathcal{T}(s_{t+1}|s_t,a_t)$, and receives reward $r_t=\mathcal{R}(s_t,a_t)$. The goal is to learn a policy $\pi_\theta$ maximizing the expected discounted return $\mathbb{E}_{\pi_\theta}\!\big[\sum_{t\geq0}\gamma^t r_t\big]$, $\gamma\in(0,1)$.

\subsection{Hierarchical Framework}
\label{sec:method_hierarchy}

\subsubsection{High-Level (\acs{HL}) Policy (Fig. \ref{fig:method_overview}-B)}
The student HL policy only observes joint proprioception in $H$ history steps:
\begin{equation}
o^{\text{trans}}_{t}
:=
\Big[
\mathbf{q}^{\text{arm}}_{t-H:t},\;
\dot{\mathbf{q}}^{\text{arm}}_{t-H:t},\;
\mathbf{a}^{\text{arm}}_{t-H-1:t-1}\Big],\; H \in \mathbb{N}
\label{eq:otrans}
\end{equation}
where $\mathbf{q}^{\text{arm}}$ and $\dot{\mathbf{q}}^{\text{arm}}$ are arm joint positions and velocities, $\mathbf{a}^{\text{arm}}$ is the previous arm joint position command.
The teacher additionally accesses privileged observation:
\begin{equation}
o^{\text{wrench}}_{t} := \big[ \mathbf{F}^{\text{xy}}_{\text{force},t},\; \Theta^{\text{yaw}}_{\text{torque},t} \big]
\label{eq:opri}
\end{equation}
where $\mathbf{F}^{\text{xy}}_{\text{force}}$ and $\Theta^{\text{yaw}}_{\text{torque}}$ are the interaction wrench applied to the payload. The \ac{HL} policy outputs both a planar base motion command for locomotion execution and arm joint actions to maintain stable and compliant transport motion with
$
\mathbf{a}^{\text{HL}}_t
=
\Big[
\mathbf{a}^{\text{LL}}_t,\;
\mathbf{a}^{\text{arm}}_t
\Big],
\mathbf{a}^{\text{LL}}_t \in \mathbb{R}^{3}, \mathbf{a}^{\text{arm}}_t \in \mathbb{R}^{6}
\label{eq:hl_action_full}
$,
where the base command $\mathbf{a}^{\text{LL}}_t$ is observed by the \ac{LL} policy:
\begin{equation}
o^{\text{cmd}}_t := [\mathbf{a}^{\text{LL}}_t] := 
\big[
\mathbf{v}^{\text{xy}}_{\text{cmd},t},\;
\omega^{\text{yaw}}_{\text{cmd},t}
\big]
\label{eq:ocmd}
\end{equation}
while $\mathbf{a}^{\text{arm}}_t$ is the target arm joint positions.

\subsubsection{Low-Level (\acs{LL}) Policy (Fig. \ref{fig:method_overview}-C)}
\label{sec:ll_obs_act}
Following prior velocity-tracking backbones for legged mobile manipulation \cite{ma2022combining, zhang2024learning}, the pre-trained \ac{LL} policy $\pi_{\text{LL}}$ robustly executes the planar command on uneven terrain, taking locomotion proprioception and the \ac{HL} base command and outputting leg joint position commands with
$
\mathbf{a}^{\text{leg}}_t
=
\pi_{\text{LL}}\!\left(o^{\text{loco}}_{t}, o^{\text{cmd}}_{t}\right)
\in \mathbb{R}^{n_{\text{leg}}}.
\label{eq:ll_policy}
$,
where the observation $o^{\text{loco}}_{t}$ contains three components,
\begin{equation}
o^{\text{loco}}_t
:=
\big[
o^{\text{prop}}_t,\;
o^{\text{pred}}_t,\;
o^{\text{scan}}_t
\big],
\label{eq:oll}
\end{equation}
where $o^{\text{prop}}_t$ includes leg proprioceptions; $o^{\text{pred}}_t$ summarizes predicted external wrenches that act on the base, the current command, and base twist; and $o^{\text{scan}}_t$ provides a local height scan around the feet. During \ac{HL} policy training, $\pi_{\text{LL}}$ is kept fixed to avoid additional optimization complexity.

\subsection{Intent Estimator in HL Student Policy}
\label{sec:method_estimator}
A key challenge in deployable intent-aware transport is that the payload interaction wrench is unobservable. Instead, partner-applied wrenches leave time-correlated signatures in joint proprioception, which is readily accessible in the real world.
We therefore augment the student HL policy with an intent estimator that infers the interaction wrench from proprioceptive histories and feeds it back into the policy observation. This estimator learns an inverse of the observation kernel $\Omega$, recovering the latent intent from the proprioceptive signals that interaction induces.
{\label{sec:method_observability}This inversion is well posed under payload coupling condition. Treating the partner's action as an external wrench $\mathbf{w}$ applied at the end-effector through the contact Jacobian $\mathbf{J}$, the dynamics of the arm joints $\mathbf{q}\!\equiv\!\mathbf{q}^{\text{arm}}$ is}
\begin{equation}
{
\mathbf{M}(\mathbf{q})\ddot{\mathbf{q}}+\mathbf{C}(\mathbf{q},\dot{\mathbf{q}})\dot{\mathbf{q}}+\mathbf{g}(\mathbf{q})=\boldsymbol{\tau}+\mathbf{J}(\mathbf{q})^{\top}\mathbf{w},}
\label{eq:arm_dyn}
\end{equation}
{where $\mathbf{M},\mathbf{C},\mathbf{g}$ are the robot's known inertial terms and the PD control torque $\boldsymbol{\tau}=\mathbf{K}_p(\mathbf{a}^{\text{arm}}-\mathbf{q})-\mathbf{K}_d\dot{\mathbf{q}}$ is determined by the joint state and the arm command, whose history is included in $o_t^{\text{trans}}$. Hence the generalized-torque residual $\mathbf{r}_t:=\mathbf{M}\ddot{\mathbf{q}}+\mathbf{C}\dot{\mathbf{q}}+\mathbf{g}-\boldsymbol{\tau}=\mathbf{J}^{\top}\mathbf{w}$ is recoverable from the proprioceptive-action history, and the planar wrench follows by the least square,}
\begin{equation}
{
\mathbf{w}=\big(\mathbf{J}\mathbf{J}^{\top}\big)^{-1}\mathbf{J}\,\mathbf{r}_t=(\mathbf{J}^{\top})^{+}\mathbf{r}_t.}
\label{eq:wrench_inv}
\end{equation}
{As the arm posture penalty in Table~\ref{tab:reward} encourage the arm to stay in a well-conditioned configuration, $\mathbf{J}^{\top}$ remains
full column rank in the planar wrench space, making $\mathbf{w}$ locally identifiable from $o^{\text{trans}}_{t}$. Instead of directly applying Eq.~\eqref{eq:wrench_inv}, which would require accurate dynamics and contact geometry, the learned temporal estimator approximates this inverse from proprioceptive histories and accounts for floating-base motion and other unmodeled effects. The derivation further predicts that longer histories improve robustness and that different wrench components induce distinct joint-level response patterns, consistent with the ablation in Sec.~\ref{sec:quantivative_single_agent} and saliency analysis in Sec.~\ref{sec:results}.}

\subsubsection{The design of the intent estimator (Fig. \ref{fig:method_overview}-B2)}
Taking the arm proprioceptive observation $o^{\text{trans}}_{t}$ as input, the intent estimator outputs an estimated interaction wrench vector 
$
\mathbf{\beta}^{\text{est}}_t :=
[\hat{\mathbf{F}}^{\text{xy}}_{\text{force},t}, \hat{\Theta}^{\text{yaw}}_{\text{torque},t}]
\label{eq:beta_est_def}
$,
which is concatenated with $o^{\text{trans}}_{t}$ as input to the actor network of the student \ac{HL} policy.
Intuitively, $\hat{\mathbf{F}}^{\text{xy}}_{\text{force},t}$ captures the translational intent component, while $\hat{\Theta}^{\text{yaw}}_{\text{torque},t}$ provides a rotational intent.

\subsubsection{Training intent estimator}
During training in simulation, we can get the ground-truth interaction wrench $o^{\text{wrench}}_{t}$.
We train the intent estimator with a regression loss:
\begin{equation}
\mathcal{L}_{\text{est}}
=
\!
\left\|
\hat{\mathbf{F}}^{\text{xy}}_{\text{force},t}
-
\mathbf{F}^{\text{xy}}_{\text{force},t}
\right\|_2^2
+
\left\|
\hat{\Theta}^{\text{yaw}}_{\text{torque},t}
-
\Theta^{\text{yaw}}_{\text{torque},t}
\right\|_2^2
\label{eq:est_loss}
\end{equation}

\subsubsection{Deployment}
At deployment, where $o^{\text{wrench}}$ is unavailable, the student relies only on the proprioceptive histories $o^{\text{trans}}_{t}$ and the estimated intent $\mathbf{\beta}^{\text{est}}_t$, retaining an explicit intent representation while remaining lightweight.

\subsection{Teacher-Student Training with KL Regularization}
\label{sec:method_ts}
We adopt a teacher-student scheme to improve deployability: a privileged teacher first learns a well-shaped intent-to-motion mapping, which is then distilled into a deployable student relying only on proprioceptive observations.

\subsubsection{Privileged teacher (Fig. \ref{fig:method_overview}-B1)}
We first train a privileged \ac{HL} teacher policy $\pi_{\text{HL}}^{\text{T}}$ using PPO, with access to privileged intent observation $o^{\text{wrench}}$. 
The teacher learns an admittance-like mapping from interaction wrenches to arm and base motion commands while remaining robust to load variations and terrain-induced disturbances introduced during training.

\subsubsection{Deployable student with KL regularization (Fig. \ref{fig:method_overview}-B3)}
We then train a student policy $\pi_{\text{HL}}^{\text{S}}$  (Fig. \ref{fig:method_overview}-B2) that only relies on proprioceptive observation $o^{\text{trans}}$, where we use a short history window $H$ to capture meaningful interaction cues, and optimize a joint objective composed of a standard expected return and a KL distillation term with
\begin{equation}
\max_{\theta_{\text{S}}}\;
\mathbb{E}\!\left[\sum_{t} \gamma^{t} r_t\right]
\;-\;\lambda_{\text{KL}}\,\mathcal{J}_{\text{KL}}(\theta_{\text{S}})
\label{eq:kl_objective_split}
\end{equation}
\begin{equation}
\mathcal{J}_{\text{KL}}(\theta_{\text{S}}) = 
\mathbb{E}\!\left[
D_{\text{KL}}\!\Big(
\pi_{\text{HL}}^{\text{S}}(\cdot|o^{\text{trans}}_{t},\beta^{\text{est}}_t)
\,\|\,
\pi_{\text{HL}}^{\text{T}}(\cdot|o^{\text{T}}_{t})
\Big)
\right]
\end{equation}
where $o^{\text{T}}_{t} : = \{o^{\text{trans}}_{t},o^{\text{wrench}}_{t} \}$.
The expected return $\mathcal{J}_{\text{RL}}$ encourages the student to improve task performance. The KL term $\mathcal{J}_{\text{KL}}$ encourages the students to imitate teacher actions by penalizing deviations between $\pi_{\text{HL}}^{\text{S}}$ and $\pi_{\text{HL}}^{\text{T}}$.
The weight $\lambda_{\text{KL}}$ balances imitation and task performance. For both teacher and student, we use an asymmetric actor-critic \cite{lambrechts2025theoretical}, conditioning the critic on additional end-effector and payload position, velocity, and acceleration to improve value estimation and reduce gradient variance.

\begin{table}[t]
\centering
\vspace{0.2cm}
\caption{Reward For Learning the Transport Policy}
\label{tab:reward_reg}
\setlength{\tabcolsep}{4pt}
\renewcommand{\arraystretch}{1.1}
\begin{tabular}{l l r}
\toprule
\textbf{Reward Term} & \textbf{Expression} & \textbf{Weight ($\times dt^\#$)} \\
\midrule

Force Tracking &
Eq. \eqref{eq:force_award} &
$-2.0\times10^{-2}$ \\

Torque Tracking &
Eq. \eqref{eq:torque_award} &
$-1.0\times10^{-2}$ \\

Payload Height &
Eq. \eqref{eq:height_award} &
$-3.0\times10^{-3}$ \\

Joint torque penalty &
$\sum_i \tau_i^2$ &
$-2.0\times10^{-9}$ \\

Joint DoF penalty &
$\sum_i \dot q_i^2 + 0.00025\,\ddot q_i^2$ &
$-2.0\times10^{-4}$ \\

Action penalty &
$\sum_i (a_{t,i}-a_{t-1,i})^2 + 0.5\,a_{t,i}^2$ &
$-2.0\times10^{-4}$ \\

Termination &
$\mathbb{I}[\text{terminated}]$ &
$-50.0\,/\,dt$ \\

Arm posture penalty &
$\left\lVert \mathbf{q}_{\text{arm}}-\mathbf{q}^{\textit{default}}_{\text{arm}}\right\rVert_2$ &
$-3.0\times10^{-3}$ \\
\bottomrule
\end{tabular}

\begin{tablenotes}
\footnotesize
\item $dt^\# = 0.02$ s, denotes the control interval.
\end{tablenotes}

\vspace{-0.3cm}
\label{tab:reward}
\end{table}

\begin{figure*}[htpb]
  \centering
  \vspace{0.1cm}

  \begin{subfigure}[t]{\textwidth}
    \centering
    \includegraphics[width=\textwidth]{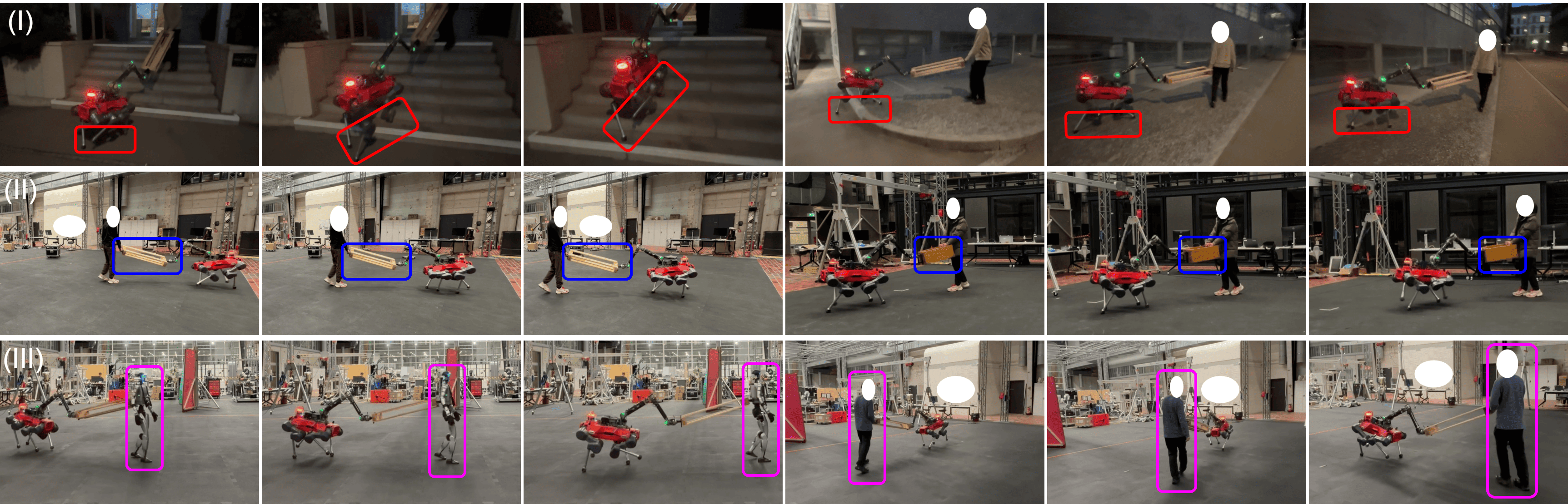}
    \vspace{-0.6cm}
    \caption{\textbf{Cooperative transport across diverse terrains, payloads, and team configurations.} {The quadruped-arm system reliably infers partner intent from arm proprioceptive histories without \ac{FT} sensors, enabling stable co-carrying (I) across uneven terrains (e.g. stairs and gravel road), (II) with different payload (e.g. short and long box), and (III) with different partners (e.g. human and humanoid).}}
    \label{fig:real_exp}
  \end{subfigure}
  \vspace{0.2cm}

  \begin{subfigure}[t]{\textwidth}
    \centering
    \includegraphics[width=\textwidth]{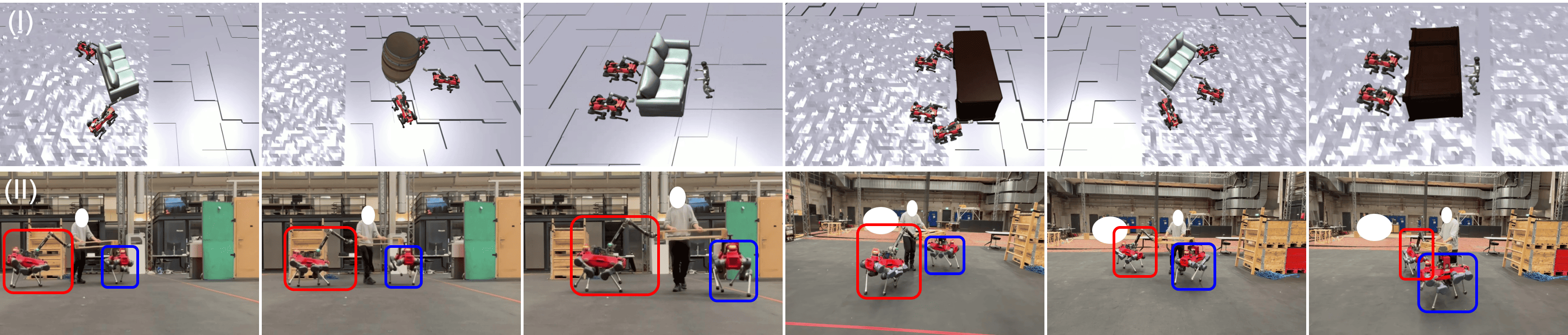}
    \vspace{-0.6cm}
    \caption{{\textbf{Cooperative transport with multi-robot teams and diverse payloads in simulation (I) and real world (II).} The proposed approach scales from pairwise co-carrying to decentralized teams of 2-4 legged robots that jointly move large, heavy, and irregular payloads.}}
    \label{fig:multi_agent}
  \end{subfigure}
  \vspace{0.2cm}

  \begin{subfigure}[t]{\textwidth}
    \centering
    \includegraphics[width=\textwidth]{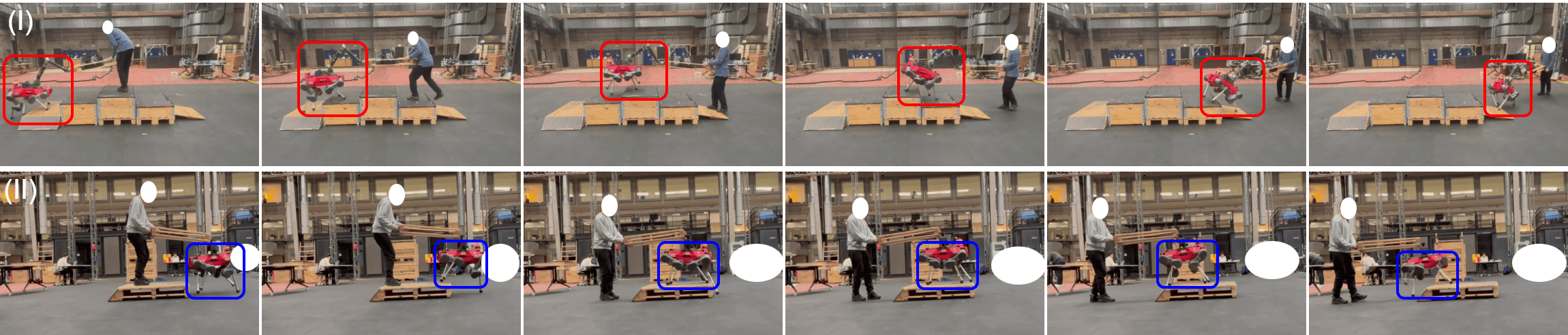}
    \vspace{-0.6cm}
    \caption{\textbf{Cooperative transport with diverse legged robots.} {The proposed high-level intent policy enables robust co-carrying with an one-arm ANYmal (I) over rough or stepping terrain, and also generalizes to an ANYmal without an arm (II) for payload transport.}}
  \label{fig:cross_embodiment}
  \end{subfigure}

  \caption{\textbf{Overview of partner-agnostic intent-aware cooperative transport results.} The proposed intent-aware framework enables stable cooperative transport across diverse terrains, payloads, partners, robot teams, and robot embodiments.}
  \label{fig:results_overview}
  \vspace{-0.4cm}
\end{figure*}

\subsection{Interaction Wrench Generation in Simulation}
\label{sec:wrench_generation}
To generate diverse and smooth interaction cues during training, we apply an interactive wrench profile (Fig. \ref{fig:method_overview}-A) to the payload.
For each episode, we sample a planar force $\mathbf{F}^{\text{xy}}_T$ and yaw moment $\Theta^{\text{yaw}}_T$, with $\mathbf{F}^{\text{xy}}_T\in\{\SI{-40}{\newton},\SI{40}{\newton}\}$ and $\Theta^{\text{yaw}}_T\in\{\SI{-10}{\newton\cdot\meter},\SI{10}{\newton\cdot\meter}\}$, and apply a time-varying scaling $s(t)\in[0,1]$,
$
\mathbf{F}^{\text{xy}}_{\text{force}}(t) = s(t)\,\mathbf{F}^{\text{xy}}_T, \Theta^{\text{yaw}}_{\text{torque}}(t) = s(t)\,\Theta^{\text{yaw}}_T,
\label{eq:wrench_profile}
$
where $s(t)$ uses a piecewise-linear schedule as follows. We set
$t_{\uparrow}=0.1T$, $t_{\text{hold}}=0.8T$, and $t_{\downarrow}=0.1T$, where $T$ denotes the duration of one episode:
\begin{equation}
s(t)=
\begin{cases}
\frac{t}{t_{\uparrow}}, & 0\le t<t_{\uparrow},\\
1, & t_{\uparrow}\le t<t_{\uparrow}+t_{\text{hold}},\\
1-\frac{t-(t_{\uparrow}+t_{\text{hold}})}{t_{\downarrow}}, & t_{\uparrow}+t_{\text{hold}}\le t<t_{\uparrow}+t_{\text{hold}}+t_{\downarrow}
\end{cases}
\label{eq:ramp_hold_ramp}
\end{equation}
The piecewise-linear schedule avoids impulsive transients and yields physically plausible interaction profiles. To encourage fully compliant behavior, we also optionally sample zero-wrench segments with a fixed probability $p = 0.02$.

\subsection{Task Objective and Reward Design}
\label{sec:reward_design}

Cooperative transport must follow the partner's intended motion while keeping the payload stable. Since the interaction wrench encodes this intent, we design rewards that align the robot base motion with the wrench on the payload.
The corresponding rewards are defined as:
\begin{equation}
r_{\text{force}} = \exp\!\Big(- \left\lVert \mathbf{F}^{\text{xy}}_{\text{force}}/B_{\text{force}}-\mathbf{v}^{\text{xy}}_{\text{base}}\right\rVert_2^2\; / \;{\sigma_{\text{force}}}\Big),
\label{eq:force_award}
\end{equation}
\begin{equation}
r_{\text{torque}} = \exp\!\Big(- \left(\Theta^{\text{yaw}}_{\text{torque}}/B_{\text{torque}}-\omega^{\text{yaw}}_{\text{base}}\right)^2\; / \;{\sigma_{\text{torque}}}\Big).
\label{eq:torque_award}
\end{equation}
Here $\mathbf{F}^{xy}_{\text{force}}$ and $\Theta^{yaw}_{\text{torque}}$ denote the interaction force and torque applied on the payload, and $\mathbf{v}^{xy}_{\text{base}}$ and $\omega^{yaw}_{\text{base}}$ are the robot base linear velocity and yaw rate. The coefficient $B_{\text{force}}$ and $B_{\text{torque}}$ scale the wrench into a velocity reference, similar to the gain in classical admittance control \cite{zhou2025hac}. Meanwhile, to ensure stable transport, we penalize payload height deviations outside a safe range with:
\begin{equation}
r_{\text{h}} =
\left(
\max(0,h_{\min}-h_{\text{base}}^{\text{p}})
+
\max(0,h_{\text{base}}^{\text{p}}-h_{\max})
\right)^2
\label{eq:height_award}
\end{equation}
where $h_{\text{base}}^{\text{p}}$ is the measured payload height. $(h_{\min}, h_{\max})$ is the safe height range. All height values are represented in the robot base frame. This term prevents excessive tilting or dropping of the payload during transport. In our implementation, we set $\sigma_{\text{force}}=0.25$, $\sigma_{\text{torque}}=0.25$, $B_{\text{force}}=40$, $B_{\text{torque}}=10$, $h_{\min}=\SI{0.05}{\meter}$ , and $h_{\max}=\SI{0.25}{\meter}$. All rewards are summarized in Table \ref{tab:reward}.

\section{Implementation Details}

We train our policies in Isaac Gym using 4096 parallel environments on a single RTX 3080Ti GPU. We additionally randomize the payload mass to avoid overfitting to a single payload-dynamics regime, sampling the mass uniformly from \SI{0}{\kilogram} to \SI{10}{\kilogram}. The LL policy follows the setup in \cite{ma2022combining}. For the HL policy, both the teacher and the student use three-layer MLP actor and critic networks with hidden dimensions (128, 128, 128). The intent estimator network adopts the same MLP architecture. The teacher and student HL policies are trained for 15k and 20k iterations. To further increase policy robustness and reduce sim-to-real gaps, domain randomization is applied to the robot's proprioceptive observations and the generated interactive wrench.

\begin{table*}[htpb]
  \centering
  \vspace{0.2cm}
  \caption{Quantitative comparison on collaborative transport metrics under different payload masses (2/4/8 kg).}
  \scriptsize
  \setlength{\tabcolsep}{1.5pt}
  \renewcommand{\arraystretch}{0.92}
  \begin{adjustbox}{max width=\textwidth}
  \begin{tabular}{lccccc|ccc}
    \toprule
    \multirow{2}{*}{Metrics} 
    & \multicolumn{5}{c}{Baseline Comparisons}
    & \multicolumn{3}{c}{Ablation Studies} \\
    \cmidrule(lr){2-6} \cmidrule(lr){7-9}
    & \multicolumn{1}{c}{Damped Arm}
    & \multicolumn{1}{c}{Pure RL-4}
    & \multicolumn{1}{c}{BC Distill.-4}
    & \multicolumn{1}{c}{Force Est.-4}
    & \multicolumn{1}{c}{\textbf{PAINT-4}}
    & \multicolumn{1}{c}{PAINT-4}
    & \multicolumn{1}{c}{\textbf{PAINT-8}}
    & \multicolumn{1}{c}{PAINT-1} \\
    \midrule

    $^\ast$Lin. Trk. Err. ($m/s$) $\downarrow$
    & 0.331/0.290/0.294
    & 0.143/0.179/0.237
    & 0.224/0.254/0.275
    & 0.138/0.148/0.225
    & \textbf{0.120}/\textbf{0.142}/\textbf{0.220}
    & 0.120/0.142/0.220
    & \textbf{0.119}/\textbf{0.135}/\textbf{0.202}
    & 0.140/0.168/0.267 \\

    $^\ast$Ang. Trk. Err. ($rad/s$) $\downarrow$
    & 0.563/0.491/0.452
    & 0.120/0.163/0.333
    & 0.364/0.373/0.451
    & 0.132/0.157/0.309
    & \textbf{0.111}/\textbf{0.119}/\textbf{0.284}
    & 0.111/\textbf{0.119}/0.284
    & \textbf{0.105}/0.150/\textbf{0.252}
    & 0.123/0.140/0.315 \\

    $^\dag$F. Est. Err. ($N$) $\downarrow$
    & --/--/--
    & --/--/--
    & --/--/--
    & 3.321/4.361/\textbf{7.875}
    & \textbf{3.041}/\textbf{4.008}/7.956
    & 3.041/4.008/7.956
    & \textbf{2.764}/\textbf{3.471}/\textbf{6.658}
    & 4.367/5.117/8.041 \\

    $^\dag$T. Est. Err. ($N{\cdot}m$) $\downarrow$
    & --/--/--
    & --/--/--
    & --/--/--
    & \textbf{0.640}/\textbf{0.921}/\textbf{2.779}
    & 0.781/0.923/2.881
    & 0.781/\textbf{0.923}/2.881
    & \textbf{0.566}/1.022/\textbf{2.591}
    & 0.972/1.159/3.387 \\

    $^\ddag$E.E. Force ($N$) $\downarrow$
    & 8.403/14.671/27.910
    & 3.351/10.791/\textbf{25.031}
    & 10.712/17.138/29.722
    & \textbf{2.269}/7.404/27.319
    & 2.929/\textbf{7.291}/25.982
    & 2.929/7.291/25.982
    & \textbf{2.024}/\textbf{7.136}/\textbf{25.962}
    & 3.663/11.213/28.746 \\

    $^\ddag$E.E. Torque ($N{\cdot}m$) $\downarrow$
    & 4.168/6.241/12.637
    & 1.605/3.617/11.566
    & 2.364/6.898/12.835
    & \textbf{1.596}/3.401/10.935
    & 1.857/\textbf{2.346}/\textbf{10.621}
    & 1.857/\textbf{2.346}/10.621
    & \textbf{1.098}/2.688/\textbf{10.055}
    & 1.672/3.720/11.278 \\

    $^\S$Intent Align. $\uparrow$
    & 0.612/0.615/0.465
    & 0.880/0.902/0.754
    & 0.785/0.736/0.557
    & 0.908/0.910/0.777
    & \textbf{0.923}/\textbf{0.915}/\textbf{0.781}
    & 0.923/0.915/0.781
    & \textbf{0.927}/\textbf{0.921}/\textbf{0.842}
    & 0.905/0.901/0.669 \\

    \bottomrule
  \end{tabular}
  \end{adjustbox}

  \begin{tablenotes}
    \scriptsize
    \item ``Trk.'', ``Est.'', and ``E.E.'' denote tracking, estimation, and end-effector, respectively.
    \item $^\ast$ The tracking errors between robot base motion and interaction force and torque, as explained in \ref{sec:reward_design}.
    \item $^\dag$ The mean absolute errors between the estimated force and torque and their ground truth.
    \item $^\ddag$ The magnitude of the force and torque exerted by the robot end-effector on the payload during steady-state transport.
    \item $^\S$ The cosine similarity between commanded force and end-effector velocity direction:
    $\mathrm{cossim}\!\left(\mathbf{F}_{\text{force},t}, \mathbf{v}_{\text{ee},t}\right)
    = \mathbf{F}_{\text{force},t}^{\top}\mathbf{v}_{\text{ee},t}/
    {\|\mathbf{F}_{\text{force},t}\|_2 \, \|\mathbf{v}_{\text{ee},t}\|_2}$.
  \end{tablenotes}
  \vspace{-0.6cm}
  \label{tab:compared_method}
\end{table*}

\begin{figure}[tpb]
  \centering
  \includegraphics[width=\linewidth]{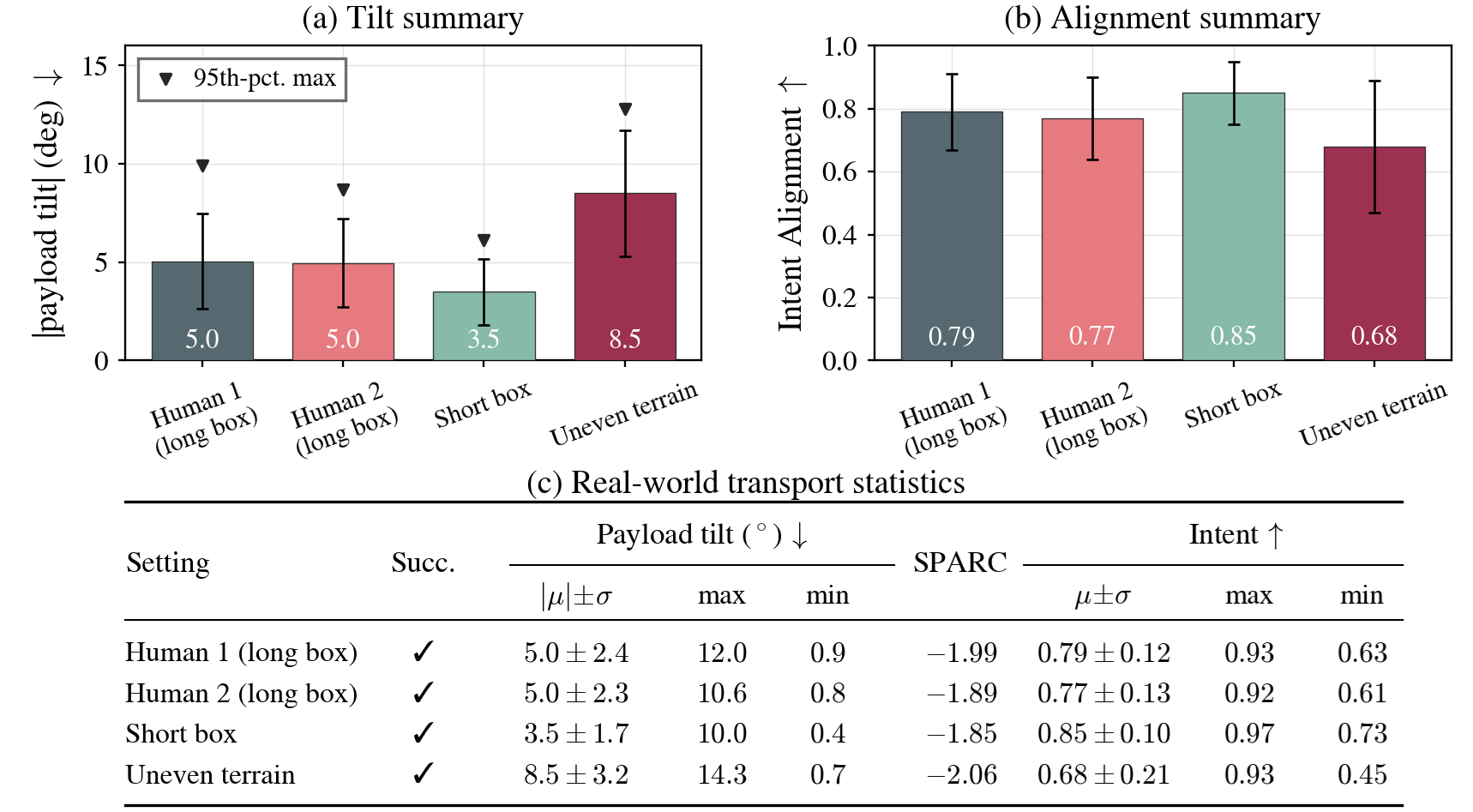}
  \vspace{-0.5cm} 
  \caption{{Real-world transport statistics across diverse leaders, payloads, and terrain. (a) The shared payload stays near level and (b) the partner-intent alignment remains high, with (c) per-condition statistics (success, payload tilt, SPARC smoothness, and intent alignment) summarized in the table.}}
  \label{fig:alma_comparison}
  \vspace{-0.2cm}
\end{figure}
\section{Experiments and Results}
\label{sec:results}

\begin{figure}[tpb]
  \centering
  \includegraphics[width=\linewidth]{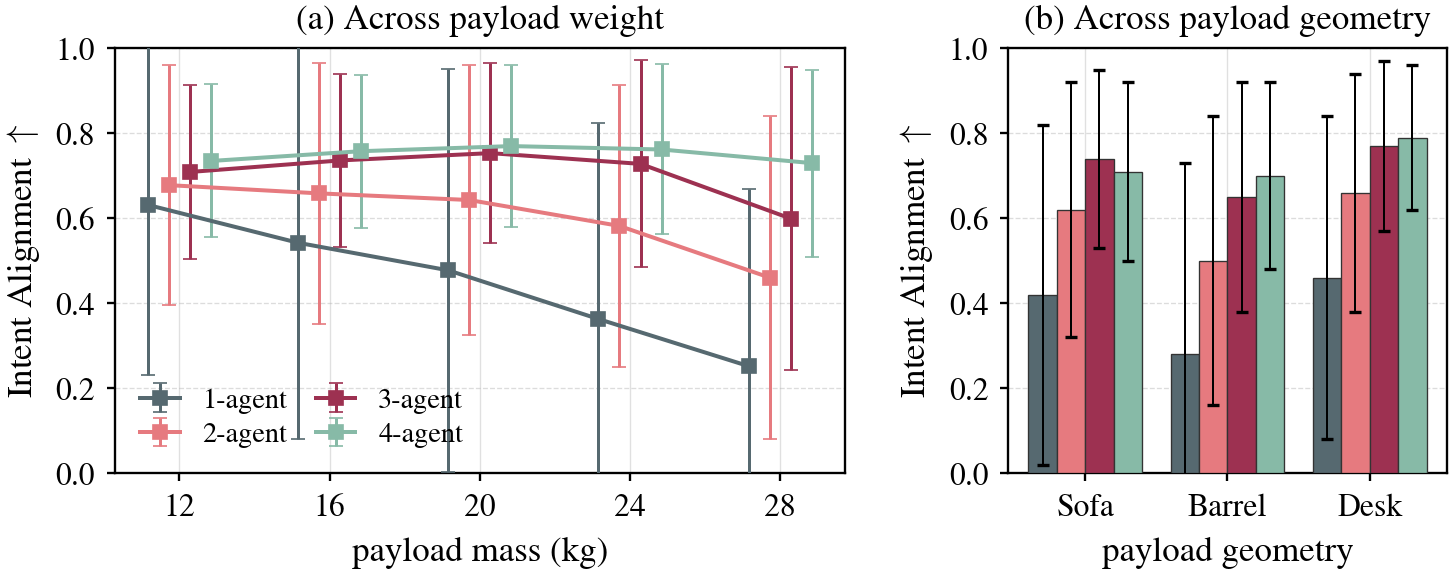}
  \vspace{-0.4cm} 
  \caption{{Simulation analysis of scalability. Partner-intent alignment is evaluated for teams of one to four robots under varying (a) payload masses ($12$-$28\,$kg) and (b) payload geometries. A single robot degrades sharply with increasing payload mass, whereas three- and four-robot teams sustain high alignment, showing improved handling of heavy payload-coupled dynamics.}}
  \label{fig:multi_agent_quantitative}
  \vspace{-0.6cm}
\end{figure}

\begin{figure}[tpb]
  \centering
  \includegraphics[width=\linewidth]{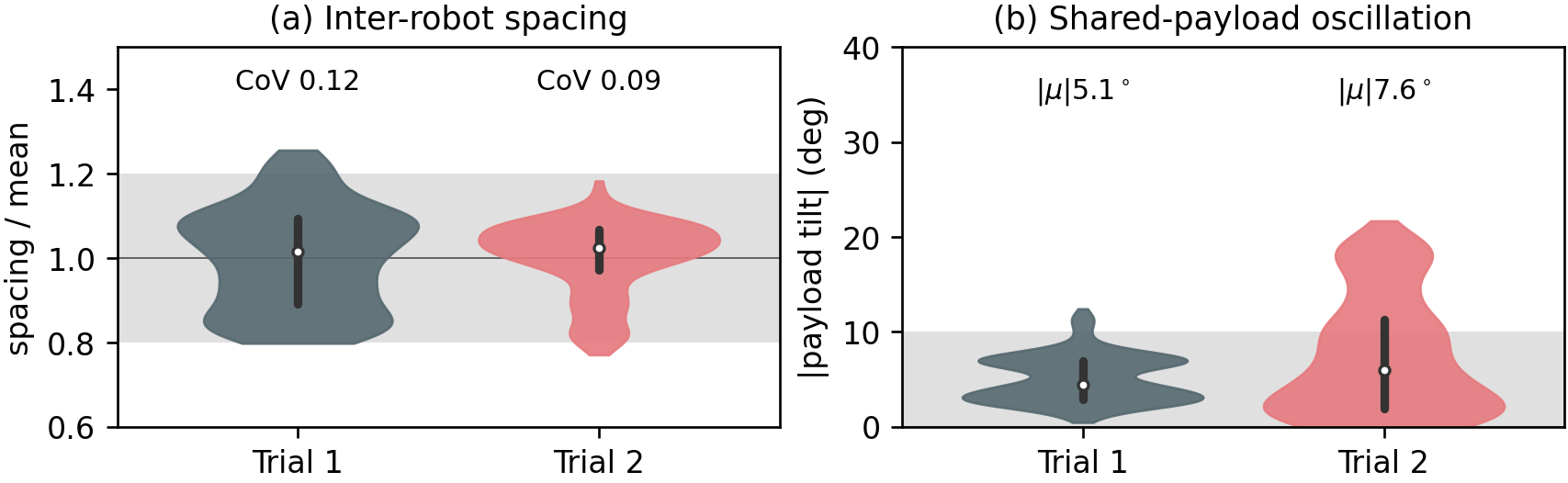}
  \vspace{-0.4cm} 
  \caption{{Real-world multi-robot coordination with one human and two robots carrying an $8$ kg long box over two repeated trials: a mostly straight (Trial 1) and a more turning-intensive (Trial 2). (a) Inter-robot spacing remains tightly distributed (CoV $\approx 0.1$), and (b) the shared payload remains near level, indicating consistent decentralized coordination through payload-coupled interaction.}}
  \label{fig:multi_agent_comparison}
  \vspace{-0.6cm}
\end{figure}

\subsection{Quantitative Studies Across Different Payload Masses}
\label{sec:quantivative_single_agent}

Table~\ref{tab:compared_method} reports the single-agent quantitative results under different payload masses and the same interaction force and torque in simulation. 
The compared baselines include Damped Arm, Pure RL-4 \cite{schulman2017proximal}, BC Distillation-4 \cite{du2025learning}, and Force Estimator-4 \cite{portela2024learning, zhi2025learning}. Damped Arm is implemented via classical PD control: 
the arm is in damping mode, and the end-effector displacement induced by external forces and torques is passed to the PD controller to compute base motion commands. A low-pass filter is used to improve motion smoothness. All other methods in baseline comparisons use 4-step (\SI{0.08}{\second}) proprioception histories, while in ablation studies, different history steps are compared.
As payload mass increases, all methods degrade, indicating more challenging coupled dynamics, yet PAINT remains the most robust, consistently achieving the best or near-best metrics. Notably, although Force Est.-4 achieves competitive \ac{FT} estimation accuracy, its tracking errors typically exceed those of PAINT-4, indicating that PAINT benefits not only from informative interaction wrenches but also from teacher-student distillation; the end-effector metrics confirm this, with PAINT producing lower constraint \ac{FT} and stronger intent alignment than the baselines. In ablation studies, we show that temporal history is critical under partial observability, where reducing history steps consistently degrades both tracking and interaction performance, whereas longer histories further improve robustness.

\subsection{Real-World Transport Across Diverse Situations}

Fig.~\ref{fig:real_exp} shows that the robot reliably infers partner intent from only arm proprioceptive histories, without \ac{FT} sensors or payload tracking. Across uneven terrains such as stairs and gravel, the HL policy produces smooth base commands that the \ac{LL} backbone tracks robustly, {and the method remains effective across payload geometries and weights and across different leaders, demonstrating partner-agnostic transport behavior.
Quantitatively, across diverse leaders and payloads, Fig.~\ref{fig:alma_comparison} demonstrates that the intent-alignment score stays high ($0.68$-$0.85$) with a small spread across leader types, and the payload stays close to level ($3.5^\circ$-$8.5^\circ$), indicating partner-agnostic intent inference and stable payload handling.}

\subsection{{Decentralized Multi-Robot Transport}}
\label{sec:results_multirobot}
Fig.~\ref{fig:multi_agent} evaluates multi-robot scalability by deploying the same transport policy on teams of 2-4 legged robots. The teams successfully transport various payloads, including a sofa, a desk, and a barrel, over uneven terrain in simulation, and jointly carry an \SI{8}{\kilogram} long box with a human leader in the real world. Although trained only in a single-agent setting with up to \SI{10}{\kilogram} payloads, PAINT remains effective on larger teams and heavier payloads. {Concretely, Fig.~\ref{fig:multi_agent_quantitative} sweeps team size, payload mass, and geometry. As mass increases, a single robot drops from $0.63$ to $0.25$ intent alignment, whereas robot teams maintain $0.70$ to $0.77$, consistent with the real-world two-robot trials in Fig.~\ref{fig:multi_agent_comparison} (spacing CoV $0.09$-$0.12$, payload tilt $|\mu|=5.1^\circ$-$7.6^\circ$). Here, zero-shot means the identical single-agent weights are reused with no fine-tuning, reward change, or observation change.} Notably, this coordination remains fully decentralized, with each robot running the same policy and inferring team-level intent from its own proprioceptive histories, without centralized planning or state sharing.
{These results are consistent with the observability argument in Sec.~\ref{sec:method_observability}: under a rigid shared payload, the leader's and teammates' wrenches are mechanically combined through the object, and each robot observes a local proprioceptive projection of this team-level interaction in the same planar intent space used by the single-agent estimator. The payload can therefore be interpreted as an implicit physical consensus channel, which helps explain the observed transfer of a single-wrench-trained controller to multi-robot teams without retraining.}

\begin{figure}[tpb]
  \centering
  \includegraphics[width=\linewidth]{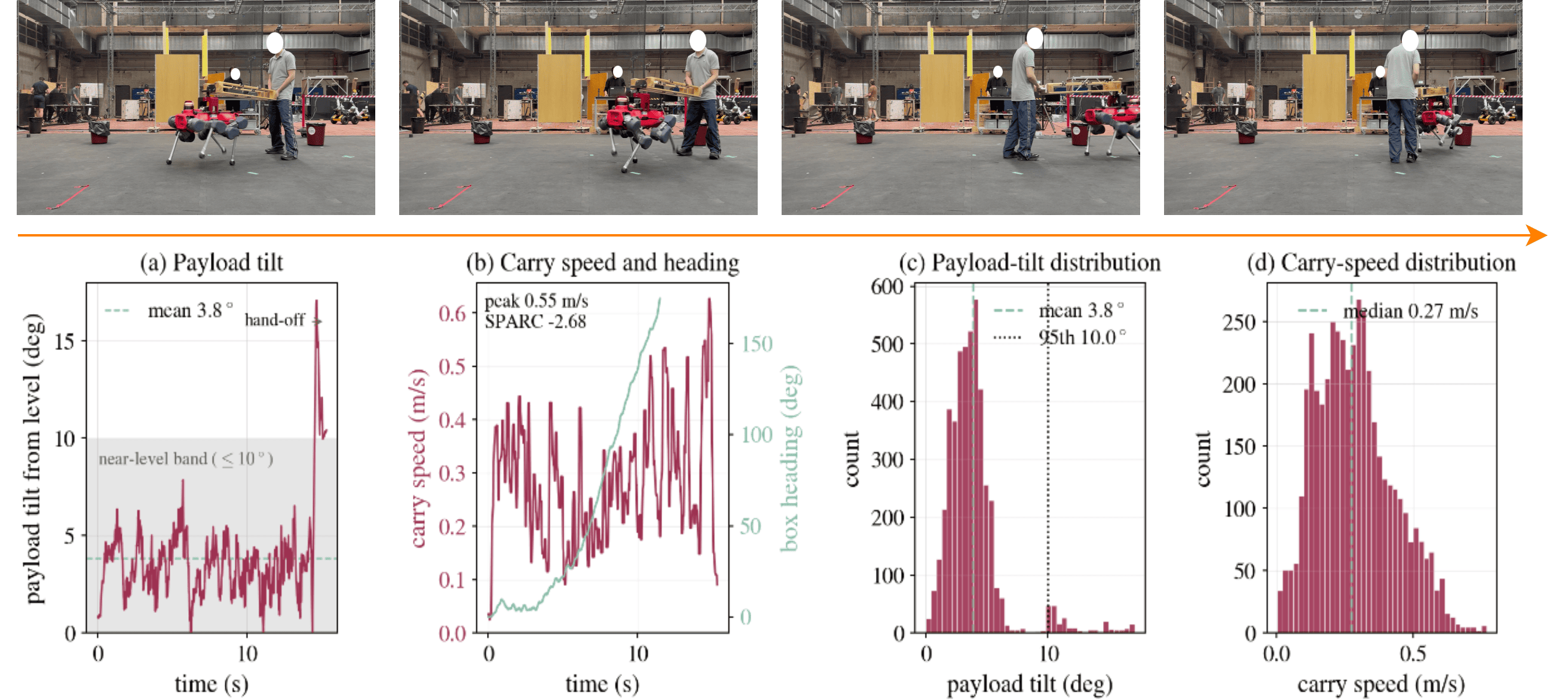}
  \vspace{-0.8\baselineskip}%
  \caption{{Real-world cross-embodiment transport with an ANYmal. While carrying a long box, the robot follows the human leader and maintains (a) small payload tilt and (b) steady carrying speed and heading, with (c) a tightly concentrated tilt distribution and (d) a concentrated carrying-speed distribution, demonstrating stable transport from proprioception alone.}}
  \label{fig:anymal_without_arm_quantitative}
  \vspace{-0.4\baselineskip}%
\end{figure}

\begin{figure}[tpb]
  \centering
  \includegraphics[scale=1.1]{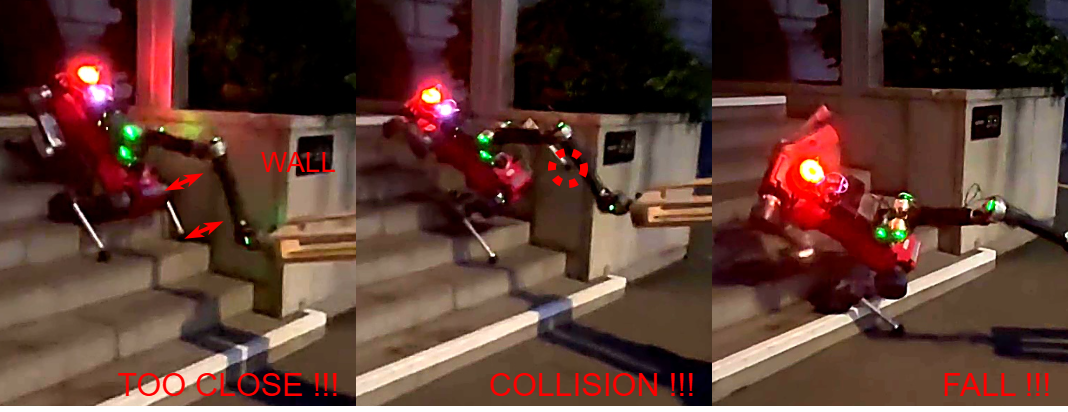}
  \caption{The \ac{HL} policy reacts purely to proprioceptive signals without active obstacle avoidance. When approaching obstacles (left), it cannot proactively adjust its motion, leading to collisions (middle) and causing locomotion failure (right).}
  \label{fig:failure_case}
  \vspace{-0.8\baselineskip}%
\end{figure}

\begin{figure*}[htpb]
  \centering
  \includegraphics[width=0.9\textwidth, trim={0 0 0 0.8cm}]{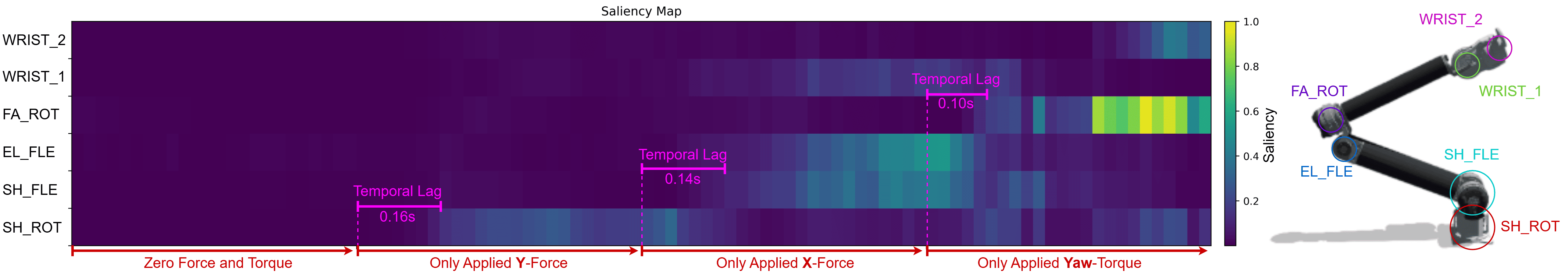}
  \caption{Saliency analysis of the intent estimator under a sequential interaction schedule across arm joint positions.}
  \label{fig:saliency}
  \vspace{-0.6cm}
\end{figure*}

\subsection{{Different Robot Embodiments Transport}}
\label{sec:results_generalization}
In this experiment, we transfer PAINT to ANYmal without an arm. For the new robot embodiment, the \ac{HL} policy takes proprioceptive signals from leg joints as input and outputs base motion commands for its \ac{LL} locomotion backbone. Fig.~\ref{fig:cross_embodiment} shows that both ANYmal with an arm and without arm achieve robust cooperative transport in simulation and in real-world trials over rough and stepping terrain. {This indicates that the wrench-structured proprioceptive formulation can be instantiated with either arm- or leg-joint proprioception, as both encode meaningful payload-coupled interaction cues.}
{Fig.~\ref{fig:anymal_without_arm_quantitative} further quantifies the real-world cross-embodiment deployment with an ANYmal. The recorded results show that the robot follows the human leader while carrying the long box, maintains low payload tilt, and achieves steady transport with median carrying speed $0.27\,\mathrm{m/s}$. These results show that PAINT maintains stable transport when instantiated with leg-joint proprioception.} The bottom row of Fig.~\ref{fig:multi_agent} further shows heterogeneous multi-robot transport, where two different robot embodiments jointly carry an \SI{8}{\kilogram} box with a human leader.

\subsection{Saliency Analysis of Intent Estimation}

To better understand how partner intent is estimated from proprioceptive signals, we perform a gradient-based saliency analysis to reveal how the intent estimator dynamically shifts its attention across individual arm joints under varying interaction modes. Specifically, we compute the gradient of the scalar mean of the estimated interaction cue $\mathbf{\beta}^{\text{est}}_t$ with respect to the transport proprioception $o^{\text{trans}}_{t}$, where a higher gradient magnitude indicates greater estimator sensitivity to that joint channel. As illustrated in Fig.~\ref{fig:saliency}, the attention patterns exhibit distinct spatial and temporal shifts depending on the applied force or torque, focusing on joint-position inputs since velocity and action inputs were less salient. A step force along the y-direction localizes saliency at the shoulder rotation joint, x-direction cues are derived from the shoulder, elbow, and wrist flexion joints, and yaw requires more combinatorial signaling led by the forearm rotation joint. We further observe a short, consistent temporal lag before salient responses emerge, suggesting the estimator integrates joint-evolution patterns before forming a stable intent estimate. {These two observations, distinct joints for distinct wrench components and a temporal integration window, empirically confirm the spatial-multiplexing and temporal-observability predictions of Sec.~\ref{sec:method_observability}, and together with the degradation under shorter histories (Table~\ref{tab:compared_method}) show that proprioceptive histories are functionally required to recover intent.}

\subsection{Failure Cases and Limitations}
\label{sec:failure_cases}

The proposed framework has limitations in environments requiring active obstacle avoidance. As shown in Fig.~\ref{fig:failure_case}, when encountering obstacles such as walls or narrow structures, the \ac{HL} policy lacks explicit perception or planning capabilities to proactively avoid collisions. Instead, it passively follows the leader's motion, sending unsafe base commands to the \ac{LL} locomotion backbone, which can cause locomotion failure when the leader does not provide sufficient corrective guidance.

\section{Conclusion and Future Work}
\label{sec:conclusion}

We present PAINT, a {partner-agnostic intent-aware framework} for cooperative transport with legged robots. By combining a terrain-robust \ac{LL} locomotion backbone with a \ac{HL} transport policy that infers partner intent purely from joint-level proprioceptive histories, our approach enables lightweight deployment without \ac{FT} sensors or payload tracking. Extensive simulation and real-world experiments demonstrate that the proposed framework supports stable cooperative transport {across diverse environments and evaluated payloads, with decentralized teams of up to four robots and two embodiment-specific policy instantiations}. Furthermore, we {show, via a saliency analysis and history-length ablation,} that proprioceptive signals induced by payload-coupled interactions can encode meaningful interaction cues, {helping address partial observability in both robot-leader and multi-robot transport teams}.

Several directions remain for future work. First, partner intent is represented by a planar wrench, which is not sufficient for more complex environments. Full 6D interaction wrenches and more semantic-aware intent representations can be explored. Second, the current \ac{HL} policy allows only limited arm motion. Future research can enable the arm to move more freely, thus improving the ability to perform payload reorientation and handle constrained maneuvers, such as turning in narrow corridors. {Third, since the hierarchical high-level policy is proprioception-based, it can be made obstacle-aware by adding exteroceptive perception and a navigation objective to its training, balancing partner-intent alignment with collision avoidance. Together with a multi-robot study of asymmetric loading, these directions would improve safety in cluttered settings and further substantiate the scalability.} Lastly, collaborative grasping and lifting can be incorporated to make the pipeline more complete and practical for payload transportation.
\section*{Acknowledgments}

We would like to thank Changan Chen, Feng Tian, Francesca Bray, and Junzhe He for their assistance with the hardware experiments. We also thank Andrei Cramariuc, Kaixian Qu and Shengzhi Wang for their thorough proofreading of the manuscript and insightful comments.


\bibliography{references}

@article{schulman2017proximal,
  title={Proximal policy optimization algorithms},
  author={Schulman, John and others},
  journal={arXiv preprint arXiv:1707.06347},
  year={2017}
}

@article{he2025attention,
  title={Attention-based map encoding for learning generalized legged locomotion},
  author={He, Junzhe and others},
  journal={Science Robotics},
  volume={10},
  number={105},
  pages={eadv3604},
  year={2025},
  publisher={American Association for the Advancement of Science}
}

@article{li2024human,
  title={Human-aware physical human--robot collaborative transportation and manipulation with multiple aerial robots},
  author={Li, Guanrui and others},
  journal={IEEE Transactions on Robotics},
  volume={41},
  pages={762--781},
  year={2024},
  publisher={IEEE}
}

@inproceedings{portela2024learning,
  title={Learning force control for legged manipulation},
  author={Portela, Tifanny and others},
  booktitle={2024 IEEE International Conference on Robotics and Automation (ICRA)},
  pages={15366--15372},
  year={2024},
  organization={IEEE}
}

@inproceedings{bethala2025h,
  title={H 2-COMPACT: Human-Humanoid Co-Manipulation via Adaptive Contact Trajectory Policies},
  author={Bethala, Geeta Chandra Raju and others},
  booktitle={2025 IEEE-RAS 24th International Conference on Humanoid Robots (Humanoids)},
  pages={1004--1011},
  year={2025},
  organization={IEEE}
}

@article{du2025learning,
  title={Learning human-humanoid coordination for collaborative object carrying},
  author={Du, Yushi and others},
  journal={arXiv preprint arXiv:2510.14293},
  year={2025}
}

@inproceedings{turrisi2024pacc,
  title={PACC: A passive-arm approach for high-payload collaborative carrying with quadruped robots using model predictive control},
  author={Turrisi, Giulio and others},
  booktitle={2024 IEEE/RSJ International Conference on Intelligent Robots and Systems (IROS)},
  pages={11139--11146},
  year={2024},
  organization={IEEE}
}

@article{zhi2025learning,
  title={Learning a unified policy for position and force control in legged loco-manipulation},
  author={Zhi, Peiyuan and others},
  journal={arXiv preprint arXiv:2505.20829},
  year={2025}
}

@article{ma2022combining,
  title={Combining learning-based locomotion policy with model-based manipulation for legged mobile manipulators},
  author={Ma, Yuntao and others},
  journal={IEEE Robotics and Automation Letters},
  volume={7},
  number={2},
  pages={2377--2384},
  year={2022},
  publisher={IEEE}
}

@inproceedings{ikeura1995variable,
  title={Variable impedance control of a robot for cooperation with a human},
  author={Ikeura, Ryojun and others},
  booktitle={Proceedings of 1995 IEEE international conference on robotics and automation},
  volume={3},
  pages={3097--3102},
  year={1995},
  organization={IEEE}
}

@article{yu2020human,
  title={Human-robot co-carrying using visual and force sensing},
  author={Yu, Xinbo and others},
  journal={IEEE Transactions on Industrial Electronics},
  volume={68},
  number={9},
  pages={8657--8666},
  year={2020},
  publisher={IEEE}
}

@article{shao2024constraint,
  title={Constraint-aware intent estimation for dynamic human-robot object co-manipulation},
  author={Shao, Yifei Simon and others},
  journal={arXiv preprint arXiv:2409.00215},
  year={2024}
}

@article{zhang2024learning,
  title={Learning to open and traverse doors with a legged manipulator},
  author={Zhang, Mike and others},
  journal={arXiv preprint arXiv:2409.04882},
  year={2024}
}

@article{pandit2025multi,
  title={Multi-quadruped cooperative object transport: Learning decentralized pinch-lift-move},
  author={Pandit, Bikram and others},
  journal={arXiv preprint arXiv:2509.14342},
  year={2025}
}

@article{wang2025shared,
  title={Shared Object Manipulation with a Team of Collaborative Quadrupeds},
  author={Wang, Shengzhi and others},
  journal={arXiv preprint arXiv:2510.00682},
  year={2025}
}

@article{an2025collaborative,
  title={Collaborative loco-manipulation for pick-and-place tasks with dynamic reward curriculum},
  author={An, Tianxu and others},
  journal={arXiv preprint arXiv:2509.13239},
  year={2025}
}

@article{lambrechts2025theoretical,
  title={A theoretical justification for asymmetric actor-critic algorithms},
  author={Lambrechts, Gaspard and others},
  journal={arXiv preprint arXiv:2501.19116},
  year={2025}
}

@article{miki2022learning,
  title={Learning robust perceptive locomotion for quadrupedal robots in the wild},
  author={Miki, Takahiro and others},
  journal={Science robotics},
  volume={7},
  number={62},
  pages={eabk2822},
  year={2022},
  publisher={American Association for the Advancement of Science}
}

@inproceedings{plotas2025control,
  title={A control scheme for collaborative object transportation between a human and a quadruped robot using the MIGHTY suction cup},
  author={Plotas, Konstantinos and others},
  booktitle={2025 IEEE International Conference on Robotics and Automation (ICRA)},
  pages={16305--16311},
  year={2025},
  organization={IEEE}
}

@article{khandelwal2025compliant,
  title={Compliant control of quadruped robots for assistive load carrying},
  author={Khandelwal, Nimesh and others},
  journal={arXiv preprint arXiv:2503.10401},
  year={2025}
}

@article{pan2025roboduet,
  title={Roboduet: Learning a cooperative policy for whole-body legged loco-manipulation},
  author={Pan, Guoping and others},
  journal={IEEE Robotics and Automation Letters},
  volume={10},
  number={5},
  pages={4564--4571},
  year={2025},
  publisher={IEEE}
}

@article{lee2020learning,
  title={Learning quadrupedal locomotion over challenging terrain},
  author={Lee, Joonho and others},
  journal={Science robotics},
  volume={5},
  number={47},
  pages={eabc5986},
  year={2020},
  publisher={American Association for the Advancement of Science}
}

@inproceedings{de2023centralized,
  title={Centralized Model Predictive Control for Collaborative Loco-Manipulation.},
  author={De Vincenti, Flavio and others},
  booktitle={Robotics: Science and Systems},
  volume={2023},
  year={2023}
}

@article{gu2025hierarchical,
  title={Hierarchical cooperative locomotion control of human and quadruped robot based on interactive force guidance},
  author={Gu, Sai and others},
  journal={IEEE/ASME Transactions on Mechatronics},
  volume={30},
  number={6},
  pages={6677--6687},
  year={2025},
  publisher={IEEE}
}

@article{schperberg2026safe,
  title={Safe whole-body loco-manipulation via combined model and learning-based control},
  author={Schperberg, Alexander and others},
  journal={arXiv preprint arXiv:2603.02443},
  year={2026}
}

@article{zhang2026cognition,
  title={Cognition to Control-Multi-Agent Learning for Human-Humanoid Collaborative Transport},
  author={Zhang, Hao and others},
  journal={arXiv preprint arXiv:2603.03768},
  year={2026}
}

@inproceedings{zhou2025hac,
  title={Hac-loco: Learning hierarchical active compliance control for quadruped locomotion under continuous external disturbances},
  author={Zhou, Xiang and others},
  booktitle={2025 IEEE/RSJ International Conference on Intelligent Robots and Systems (IROS)},
  pages={10649--10655},
  year={2025},
  organization={IEEE}
}
\bibliographystyle{IEEEtran}

\newpage

\vfill

\end{document}